\documentclass[lettersize,journal]{IEEEtran}
\usepackage{amsmath,amsfonts,bm}
\usepackage{algorithmic}
\usepackage{algorithm}
\usepackage{array}
\usepackage{bbding}
\usepackage[table]{xcolor}
\usepackage[caption=false,font=normalsize,labelfont=sf,textfont=sf]{subfig}
\usepackage{textcomp}
\usepackage{stfloats}
\usepackage{url}
\usepackage{verbatim}
\usepackage{wasysym}
\usepackage{pifont}
\usepackage{cancel} 
\usepackage{graphicx}
\usepackage{amsmath}
\usepackage{amssymb}
\usepackage{booktabs}
\usepackage{multirow}
\usepackage{threeparttable}
\usepackage{bm} 
\usepackage[colorlinks,linkcolor=red,anchorcolor=green,citecolor=blue]{hyperref}
\graphicspath{{figures/}}
\newcommand*\ie{\textit{i.e.}}
\newcommand*\eg{\textit{e.g.}}

\definecolor{myred}{RGB}{255,178,178}
\definecolor{myorange}{RGB}{255,217,178}
\definecolor{myyellow}{RGB}{255,255,178}

\begin{document}
%
\title{Triple Point Masking}

\author{Jiaming Liu, Linghe Kong,~\IEEEmembership{Senior Member,~IEEE}, Yue Wu,~\IEEEmembership{Senior Member,~IEEE}, Maoguo Gong,~\IEEEmembership{Fellow,~IEEE}, Hao Li,~\IEEEmembership{Member,~IEEE}, Qiguang Miao,~\IEEEmembership{Senior Member,~IEEE}, Wenping Ma,~\IEEEmembership{Senior Member,~IEEE}, Can Qin
	
\thanks{This work was supported by the National Natural Science Foundation of China (62276200, 62036006), the Fundamental Research Funds for the Central Universities and the CAAI-Huawei MindSpore Academic Open Fund.}
\thanks{\quad Jiaming~Liu and Linghe~Kong are with the School of Electronic Information and Electrical Engineering, Shanghai Jiao Tong University, Shanghai 200240, China. (E-mail: jmliu99@sjtu.edu.cn, linghe.kong@sjtu.edu.cn)}
\thanks{\quad Yue~Wu and Qiguang~Miao are with the School of Computer Science and Technology, Key Laboratory of Collaborative Intelligence Systems, Ministry of Education, Xidian University, Xi’an 710071, China. (E-mail: ywu@xidian.edu.cn, qgmiao@xidian.edu.cn)}
\thanks{\quad Maoguo~Gong and Hao~Li are with the School of Electronic Engineering, Key Laboratory of Collaborative Intelligence Systems, Ministry of Education, Xidian University, Xi’an, 710071, China. Maoguo~Gong is also with the College of Mathematics Science, Inner Mongolia Normal University, Hohhot 010022, China. (E-mail: gong@ieee.org, haoli@xidian.edu.cn)}
\thanks{Wenping~Ma is with the School of Artificial Intelligence, Key Laboratory of Intelligent Perception and Image Understanding of Ministry of Education, Xidian University, Xi’an 710071, China. (E-mail: wpma@mail.xidian.edu.cn)}
\thanks{Can~Qin is with the Department of Electrical and Computer Engineering, Northeastern University, Boston, MA, 02115. (E-mail: qin.ca@northeastern.edu)}
\thanks{\quad Corresponding authors: Yue Wu. (E-mail: ywu@xidian.edu.cn)}
}


\markboth{}
{Liu \MakeLowercase{\textit{et al.}}: Triple Point Masking}

\maketitle

\begin{abstract}
	Existing 3D mask learning methods encounter performance bottlenecks under limited data, and our objective is to overcome this limitation. In this paper, we introduce a triple point masking scheme, named TPM, which serves as a scalable plug-and-play framework for MAE pre-training to achieve multi-mask learning for 3D point clouds. Specifically, we augment the baseline methods with two additional mask choices (\ie, medium mask and low mask) as our core insight is that the recovery process of an object can manifest in diverse ways. Previous high-masking schemes focus on capturing the global representation information but lack fine-grained recovery capabilities, so that the generated pre-training weights tend to play a limited role in the fine-tuning process. With the support of the proposed TPM, current methods can exhibit more flexible and accurate completion capabilities, enabling the potential autoencoder in the pre-training stage to consider multiple representations of a single 3D point cloud object. In addition, during the fine-tuning stage, an SVM-guided weight selection module is proposed to fill the encoder parameters for downstream networks with the optimal weight, maximizing linear accuracy and facilitating the acquisition of intricate representations for new objects. Extensive experimental results and theoretical analysis show that five baselines equipped with the proposed TPM achieve comprehensive performance improvements on various downstream tasks. Our code and models are available at \href{https://github.com/liujia99/TPM}{https://github.com/liujia99/TPM}.
\end{abstract}

\section{Introduction}
\IEEEPARstart{A}{s} a recent self-supervised learning scheme, masked autoencoder (MAE) has shown promising applications on various modalities. Given the considerable success of MAE in natural language processing \cite{devlin2018bert,radford2019language,brown2020language} and image analysis \cite{he2022masked,feichtenhofer2022masked,chen2024context}, researchers are increasing their focus toward its application in 3D point clouds. The task holds particular significance due to the prevalence and authenticity of easily captured point clouds in the real world. Simultaneously, the massiveness and complexity of the point clouds pose challenges without annotation.
\begin{figure}[t]
	\centering
	\includegraphics[width=\linewidth]{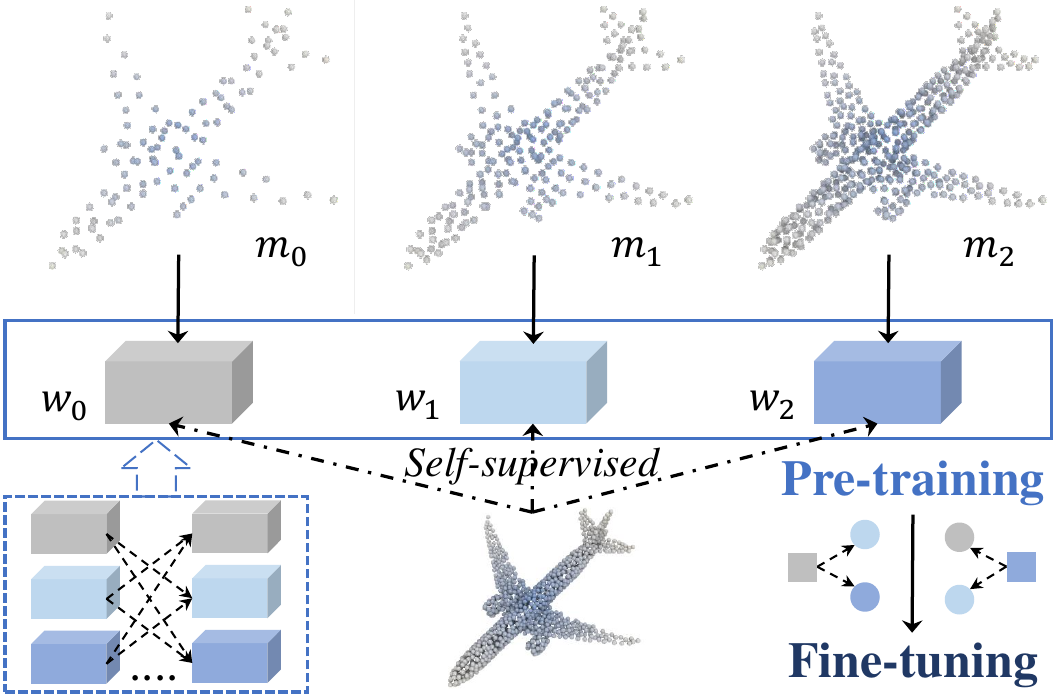}
	\caption{Illustration of TPM. Given additional masks $m_1$ and $m_2$, multi-mask completion is performed under supervision of the same input (\ie, ground truth) during pre-training. The resulting optimal weight $w_0$ or $w_1$ or $w_2$ is adopted to fit the specific encoder, providing discriminative prior conditions for downstream tasks such as classification and segmentation, etc.}
	\label{fig1}
\end{figure}

Autoencoder-based self-supervised methods \cite{yu2022point,pang2022masked,liu2022masked,zhang2022point,dong2022autoencoders,zhang2023learning,liu2023inter,yan2023implicit,jiang2023masked,chen2024pointgpt} for point clouds typically takes point patches as tokens and masks a high proportion of tokens (60\%$\sim$90\%) on the pre-training data. We observe a consistent trend, where regardless of the variations in masking techniques, autoencoder designs, and task heads, the masking ratio tends to be set at a high level. This aligns with the intuition for point cloud completion, suggesting that a higher masking ratio creates a more complex and meaningful pretext task. We may further hypothesize that completing objects with low masking rates during pre-training yield more accurate but less generalizable effects since only a small part of the point cloud can be perceived, ultimately leading to suboptimal performance in downstream tasks. Based on these analyses, we pose a question: \textit{Is it possible that existing 3D point cloud pre-training architectures be designed with multiple masking tasks to balance the advantages of each so that richer 3D representations can be obtained?}

In this paper, we present a plug-and-play architecture known as the Triple Point Masking (TPM) designed for existing 3D pre-training frameworks, as illustrated in Figure \ref{fig1}. Specifically, we integrate two additional masking choices, \ie, medium mask and low mask, for the single input. The former is introduced to balance the potential confidence bias of the other two extreme masks, while the latter offers a simple and fine-grained pre-training task. This training process is incremental, necessitating the inclusion of two extra objective functions to jointly constrain point cloud completion in different scenarios. As the learning processes for triple masking share weights, they complement each other seamlessly and do not cause additional burden. Leveraging the diversity of existing single-mask methods, including point mask expansions \cite{pang2022masked,liu2022masked}, network architecture expansions \cite{zhang2022point,yan2023implicit}, and input modality expansions \cite{dong2022autoencoders,zhang2023learning,wang2023take,liu2023inter}, our method can be easily integrated into baselines to significantly promote self-supervised learning on point clouds.
\begin{figure}
	\centering
	\includegraphics[width=\linewidth]{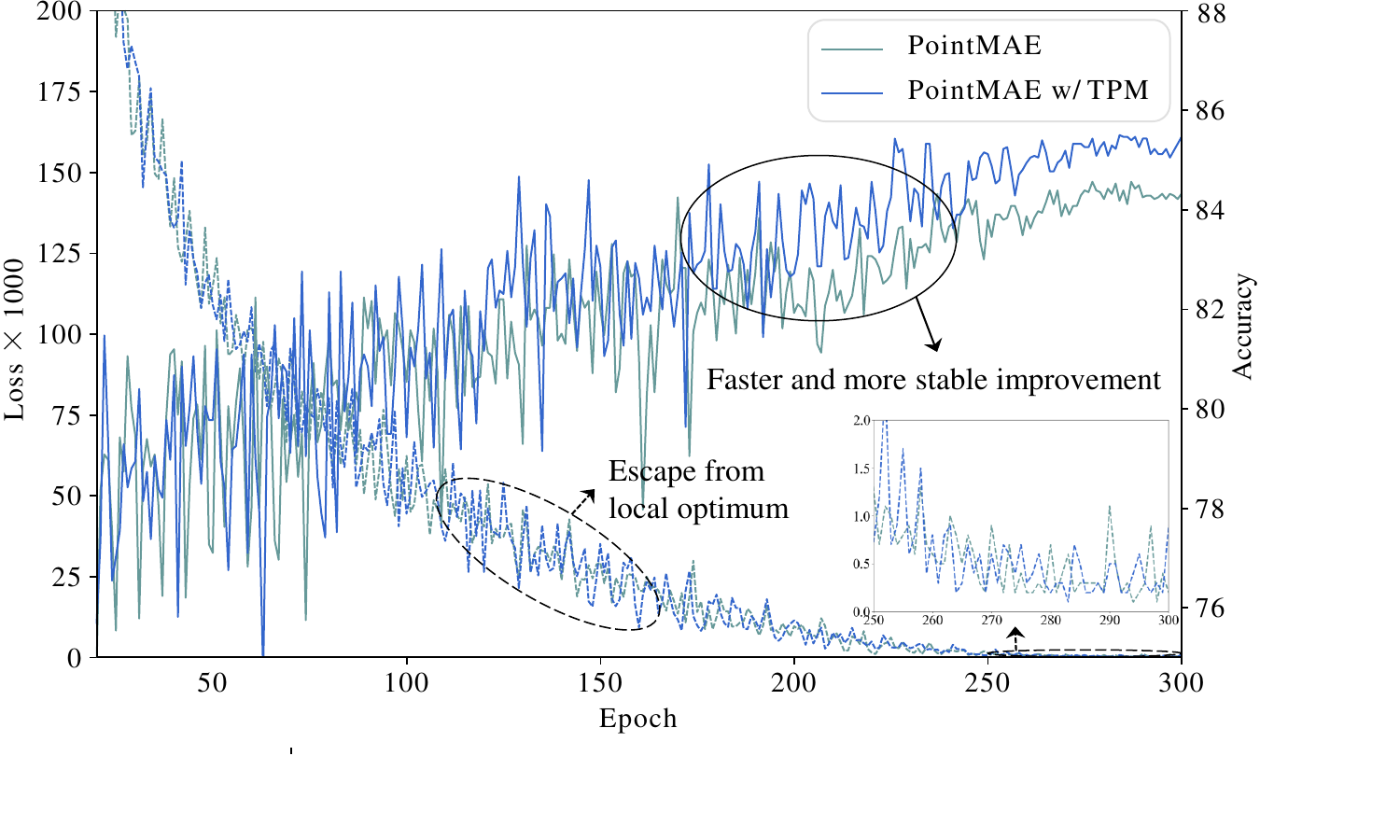}
	\caption{Comparison of training loss (left) and inference accuracy (right) of the orginal and the proposed $w_0 \mapsto {m_0}$ during fine-tuning. Results on ScanObjectNN (PB\_T50\_RS) \cite{uy2019revisiting} are reported.}
	\label{fig2}
\end{figure}

Differing from the new masks ($m_1$ and $m_2$), the vanilla mask $m_0$ acts to restore the overall spatial position of the object. However, potential errors may arise within the generated $w_0$ due to reliance solely on distance loss, leading to issues such as cross-completion between point patches and outliers being misinterpreted as inliers. As illustrated in Figure \ref{fig2}, when the vanilla $w_0$ acts on $m_0$, the training of downstream tasks becomes susceptible to overfitting and local optima, making it difficult to learn new representations. To address the issue of selecting weights based on single distance loss, we introduce an SVM-guided weight selection module to transfer the high-mask weights, which are trained with superior linear classification capabilities, thereby simulating a more discriminative effect.

Procedurally, we first preserve the optimal weight models ($w_0^*$, $w_1^*$, $w_2^*$) for the triple masks through linear support vector machine (SVM) during pre-training. Note that $w_0$, $w_1$, and $w_2$ all represent weights of the same autoencoder network, representing specific forms generated by training at different epochs of $w_{0,1,2}$. Guided by the SVM weight selection, we choose to utilize $w_0^*$ with the best linear accuracy as the only pretrained model. On the one hand, it is consistent with past fine-tuning paradigms that follow the most meaningful weight $w_0$ is the outcome of the most challenging task (\ie, high masking) during pre-training. On the other hand, this is in line with the basic principle of self-supervised learning, where a well-designed learning paradigm should efficiently initialize the network weights for subsequent fine-tuning in order to avoid weak local minima and improve stability\cite{erhan2010does}. As weight $w_0$ serves not only as a primary contributor to mask completion but also learns recovery regularities from other mask situations.

With the above multi-mask guidance and weight selection operations, our TPM can optimize the convergence of existing self-supervised methods and demonstrate significant performance improvement on various tasks. To showcase the generality of the proposed TPM, we integrate it to existing methods, including the foundational Point-MAE \cite{pang2022masked}, the modality-enhanced Inter-MAE \cite{liu2023inter}, the network-enhanced Point-M2AE \cite{zhang2022point},  the mask-enhanced PointGPT-S \cite{chen2024pointgpt} and  the data-enhanced PointGPT-B \cite{chen2024pointgpt}. Without any bells and whistles, the TPM-equipped self-supervised methods exhibit the capability to learn more robust 3D representations without changing the original conditions.

In brief, our contributions are summarized as follows:
\begin{itemize}
	\item A plug-and-play TPM module is proposed that utilizes existing 3D pre-training frameworks to learn in-depth 3D representations through triple mask completions.
	\item A weight selection strategy is introduced to create more meaningful initial conditions for the fine-tuning network to avoid overfitting problem.
	\item A series of experiments prove the importance of the proposed TPM, which remains simple yet efficient no matter how complex the original masking methods are.
\end{itemize}

\section{Related Work}
\subsection{Pre-training by Masked Autoencoders}
Masked autoencoder (MAE) can generally be divided into two steps: 1) the encoder takes randomly masked elements as input and is responsible for extracting its high-level latent representation; 2) the lightweight decoder explores clues from the encoded visible features, and reconstructs the original masked elements. Since this process only occurs in the input itself and cannot directly act on the actual function, it exists in a pre-training manner and uses the network model generated to act on other tasks. The GPTs \cite{radford2018improving,radford2019language,brown2020language} and MAEs \cite{he2022masked,liu2023mixmae,chen2024context} series have transformed this paradigm and applied it to language and image modeling, achieving significant performance improvements on downstream tasks through fine-tuning. GPT \cite{radford2018improving} adopts a unidirectional transformer architecture to fine-tune the model by updating all pre-trained parameters to implement an autoregressive prediction method. MAE \cite{he2022masked} randomly masks input patches and pre-trains the model to recover the masked patches in pixel space. In the field of 3D point clouds, Point-MAE \cite{pang2022masked} extends MAE by randomly masking point patches and reconstructing the masked regions.

\subsection{Self-supervised Learning (SSL) for Point Clouds}
With the recent emergence of zero-shot and few-shot techniques associated to data \cite{sharma2020self,zhao2021few,cheraghian2022zero,lu2023see,liu2023exploring}, self-supervised and weakly-supervised techniques related to annotations have also attracted attention. The disordered and discrete nature of 3D point clouds poses unique challenges for representation learning, so designing self-supervised solutions for point clouds is a meaningful endeavor. Different from previous mainstream constrastive learning methods \cite{sauder2019self,xie2020pointcontrast,hou2021exploring,afham2022crosspoint,wu2023self}, recent mask learning has generated multiple solutions for 3D MAEs via autoencoder structures. Point-BERT \cite{yu2022point} and Point-MAE \cite{pang2022masked} implement BERT-style \cite{devlin2018bert} and MAE-style \cite{he2022masked} point cloud pre-training schemes, respectively. MaskPoint \cite{liu2022masked} represents a point cloud as discrete occupancy values and performs a simple binary classification between masked and noisy points as an agent task. ACT \cite{dong2022autoencoders} employs a cross-modal autoencoder as a teacher model to acquire knowledge from other modalities. Point-M2AE \cite{zhang2022point} proposes a hierarchical transformer structure and a multiscale masking strategy based on Point-MAE. I2P-MAE \cite{zhang2023learning} learns excellent 3D representations from 2D pre-trained models through an image-to-point masked autoencoder. IAE \cite{yan2023implicit} adopts an implicit decoder to replace the commonly used auto encoder for better learning of point cloud representations. TAP \cite{wang2023take} proposes a point cloud-to-image generative pre-training method that generates view images with different indicated poses as a pre-training scheme through a cross-attention mechanism. PointGPS \cite{chen2024pointgpt} proposes a point cloud autoregressive generation task to pre-train the transformer model. Unlike previous MAE methods that use a standard single mask, we propose a triple mask structure and a weight selection module to re-upgrade the pre-training and fine-tuning phases of the self-supervised learning to better learn rich and robust representations for the 3D point clouds.

\subsection{Scalable SSL for Point Clouds}
Unlike expanding input data, network size, etc. on SSL with a single mask, scalable self-supervised learning for point clouds can theoretically improve the feature representation and generalisation capabilities of a model by handling tasks such as multiple point cloud representations (\eg, multiple views or multiple deformations), multiple mask learning, or multiple contrastive learning. MM-Point \cite{yu2024mm} is driven by both intra- and inter-modal similarity goals, providing multimodal interaction and transfer between multiple 2D views for a single 3D object to efficiently and simultaneously achieve coherent cross-modal learning. TriCI \cite{shao2024trici} introduces a three-branch contrast learning architecture with both within-branch and cross branch comparative learning. Each branch is equipped with a different encoder to collectively extract invariant features from different data augmentations, and features from different encoders are aligned to produce complementary and enriched learning signals. 

In contrast to appealing approaches, to the best of our knowledge our TPM is the first to set up an scalable mechanism for point masking, without the need to render 2D images corresponding to 3D point clouds or to equip multiple autoencoders. As a streamlined component, it can help existing self-supervised methods to achieve more competitive performance.

\section{Proposed Methodology}
Our objective is to design a concise and effective general-purpose component for self-supervised learning of point clouds that further facilitates the representation learning from existing methods. We first give the problem statement in Section \ref{sec3.1}. Then, we propose triple point masking and SVM-guided weight selection in Sections \ref{sec3.2} and \ref{sec3.3}, respectively. Eventually, Section \ref{sec3.4} provides baseline methods to be integrated in order to successfully deploy the proposed TPM on available self-supervised methods.
\begin{figure*}
	\centering
	\includegraphics[width=\textwidth]{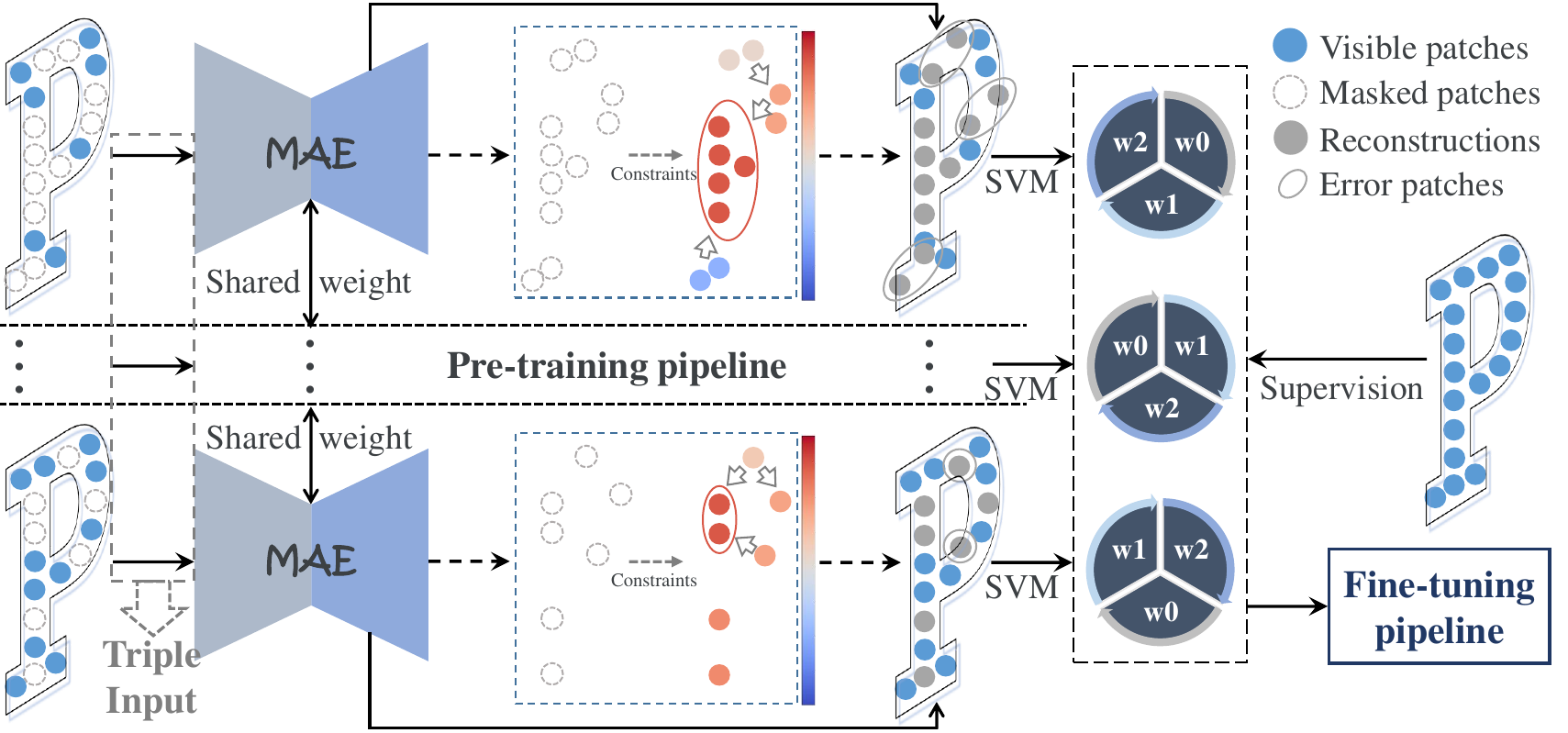}
	\caption{Overall pipeline of our TPM. Given triple masked point clouds, we extend the use of an autoencoder with shared weights corresponding to the number of inputs based on a pre-training framework (\eg Point-MAE \cite{pang2022masked}) . The autoencoder learns the recovery process under triple masks and records the respective optimal pretrained models. Being supervised by the same objective, triple mask learning can influence their respective weights for subsequently performing the weight selection operation during the fine-tuning phase.}
	\label{fig3}
\end{figure*}

\subsection{Problem Statement}\label{sec3.1}
Completion-based self-supervised learning typically starts by masking a large proportion of the inputs of a large dataset, then recovers the complete input through a small portion of the visible inputs with the help of an auto-encoder, and applies the resulting encoder model to the target dataset in order to conduct various downstream tasks. In the 3D point cloud situation, the autoencoder aims to train the encoder network $f_{\Theta}$ for extraction and the decoder network $g_{\Phi}$ for generation, where the encoder maps the input point cloud to an $c$-dimensional feature space,
\begin{equation}\label{eq1}
	f_{\Theta}: \mathbb{R}^{n \times 3} \rightarrow \mathbb{R}^c, c \ll n,
\end{equation}
where $n$ is the number of input points. Thereafter, the decoder maps the potential information in the feature space back to 3D coordinates,
\begin{equation}\label{eq2}
	g_{\Phi}: \mathbb{R}^c \rightarrow \mathbb{R}^{\tilde{n} \times 3}, \tilde{n} \leq n,
\end{equation}
where $\tilde{n} = n$ is generally set to refine the recovery process.

During the autoencoder training, the parameters within $\Theta$ and $\Phi$ are jointly trained by minimizing the distance metric (\eg, Chamfer distance or Earth Mover's distance \cite{fan2017point}) between the input and the reconstructed point cloud,
\begin{equation}
	\Theta^*, \Phi^*=\underset{\Theta, \Phi}{\arg \min}\ d(\mathcal{P}^{g_{\Phi} \circ f_{\Theta}}, \mathcal{P}),
\end{equation}
where $\mathcal{P}$ is the input point cloud and $\mathcal{P}^{g_{\Phi} \circ f_{\Theta}}$ is the reconstructed point cloud after the action of the extractor and generator. After the autoencoder is trained, the encoder $f_{\Theta}$ is fine-tuned on small target datasets with task-specific annotations (\eg, classfication and segmentation labels).

\subsection{Triple Point Masking}\label{sec3.2}
We conclude from existing research that self-supervised learning of point clouds is still affected by data sources, such as problems of unbalanced data densities, unstable sampling transformations, and limited supervised signals. Indeed, there may be an infinite number of representations of the same point cloud and an infinite number of ways in which it can be reconstructed, so that scrutinizing the problem from a geometric space infers that the point cloud always contains a unique defect \cite{yan2023implicit}. Such defects are forced to be learned by the encoder and, being subject to an overall distance metric, another point cloud generated by the decoder is forced to be identical to the input sample. Further, under high masking, the learning of the autoencoder may encounter more complex completion tasks.

Instead of indirectly turning explicit points into implicit representations \cite{yan2023implicit} or adding additional representations to them \cite{zhang2023learning}, we directly delve into the mechanism of masking and propose a multi-mask solution called triple point masking (TPM), as shown in Figure \ref{fig3}. The proposed TPM not only alleviates the singularity and complexity of the previous original pre-training task and adds multiple learning paradigms to prevent ambiguity. In other words, our TPM enables the original pre-training network to learn triple adaptive representations of the point cloud under multiple different constraints and build a reliable reconstruction pattern for the input point cloud.

Specifically, we impose two new masks $m_1$ and $m_2$ ($m_0 > m_1 > m_2$, see Table \ref{tab6} for more constructions) on top of the original $m_0$. Generally $m_0 > 0.5$ is the only setting in the baseline approach, and the two new masks we provide enable the network to mine more fine-grained information while maintaining training stability and convergence (see Figure \ref{fig2}). Based on Equations \ref{eq1} and \ref{eq2}, TPM can be mathematically expressed as
\begin{equation}\label{eq4}
	\texttt{TPM}: \left\{\begin{array}{cc}
		f_{\Theta}^{m_i}: \mathbb{R}^{n \times 3} \rightarrow \mathbb{R}^c,\\
		g_{\Phi}^{m_i}: \mathbb{R}^c \rightarrow \mathbb{R}^{\tilde{n} \times 3},	\end{array}\right. i = 0, 1, 2.
\end{equation}

Even though two autoencoders are expanded, they are still supervised by the same complete input point cloud and therefore produce their own optimal distances from different reconstructed point clouds,
\begin{equation}
	\Theta^*_i, \Phi^*_i=\underset{\Theta_i, \Phi_i}{\arg \min}\ d(\mathcal{P}^{g_{\Phi}^{m_i} \circ f_{\Theta}^{m_i}}, \mathcal{P}).
\end{equation}

Notably, considering the different difficulties encountered during the recovery process for point clouds with different masks, we set a larger loss weight for higher mask rates. As shown in Figure \ref{fig3}, our intuition is that the high-mask completion (top) is subject to more and more complex constraints (\eg, greater sparsity and more biased discreteness) than the low-mask completion (bottom) so that high-mask recovery is more hindered. Therefore, the autoencoder network should focus on the overall recovery of the point cloud while taking into account the fine-grained ones. In order to achieve this criterion, we set the weight of the $i$-th triplet loss $\mathcal{L}_{m_i}$ is $\mathcal{\lambda}_{m_i}=\frac{m_i}{\sum_{j=0}^3 m_j}$.

\subsection{SVM-Guided Weight Selection}\label{sec3.3}
Since the proposed TPM is subject to the joint action of masks $\{m_0,m_1,m_2\}$, weights $\{w_{0,1,2}^{e_i}\} (1 \le {e_i} \le \mathbf{E})$ are generated during the pre-training process, where $\mathbf{E}$ represents the number of epochs. In order to make sense of the initial conditions of the fine-tuning networks and to achieve a lightweight deployment, we select the appropriate pre-trained weights only among the $\{w_0^{e_i}\}$ generated in the toughest mask case. According to the experience of previous work \cite{pang2022masked,chen2024pointgpt}, determining the optimal weights by loss value is a straightforward strategy. However, this does not meet the needs of our design, as our losses are generated by triple mask tasks and there are differences in loss weights across tasks, making the single-masked loss an insufficient measurement of the weighting model.

As a simple and effective solution, we directly evaluate the weights $\{w_0^{e_i}\}$ by a linear SVM, and select $w_0^*$ with the maximum linear classification accuracy, \ie,
\begin{equation}
	w_0^* = \underset{w_0^{^{e_i}}}{\arg \max}\ {SVM}(\mathcal{D}_{val}|\mathcal{D}_{train}),
\end{equation}

where $\mathcal{D}_{train}$ and $\mathcal{D}_{val}$ are the data partitioned for SVM, which is chosen for ModelNet40 \cite{wu20153d} in our experiments. 

Since the linear SVM can solve maximum margin hyperplanes in linearly differentiable problems, it can be transformed into an equivalent quadratic convex optimization process. Furthermore, the weights $\{w_0^{e_i}\}$ can be quantized to discriminate the high-dimensional feature space of the point cloud under the guidance of SVM.
\begin{table*}[htb]
	\centering
	\caption{Pre-training hyperparameter settings for four baseline methods with or without TPM, where only the masks are changed.}
	\label{tab1}
	\resizebox{\linewidth}{!}{
		\setlength\tabcolsep{1mm}{
		\begin{threeparttable}	
			\begin{tabular}{l|c|c|cc|ccc|ccc}
				\toprule
				\multirow{2}*{Method} & \multirow{2}*{Input} & \multirow{2}*{Mask} &\multicolumn{2}{c|}{Patch} & \multicolumn{3}{c|}{Encoder} & \multicolumn{3}{c}{Decoder}\\	
				&  &  & Number & Size & Dimension & Depth & Head & Dimension & Depth & Head\\			
				\midrule
				Point-MAE \cite{pang2022masked} & 1024 & 0.6 $\rightarrow$ \textbf{[0.6, 0.5, 0.4]} & 64 & 32  & 384 & 12 & 6& 384 & 4 & 6\\
				Inter-MAE \cite{liu2023inter} & P+V\tnote{$\dag$} & 0.6 $\rightarrow$ \textbf{[0.6, 0.5, 0.4]} & 64 & 32  & 384 & 12 & 6& 384 & 4 & 6\\
				PointGPT-S/B \cite{chen2024pointgpt} & 1024 & 0.7 $\rightarrow$ \textbf{[0.7, 0.5, 0.3]} & 64 & 32  & 384(S)/768(B) & 12 & 6(S)/12(B) & 384(S)/768(B) & 4 & 6(S)/12(B)\\
				Point-M2AE \cite{zhang2022point} & 2048 & 0.8 $\rightarrow$ \textbf{[0.8, 0.5, 0.2]} & [512, 256, 64] & [16, 8, 8]  & [96, 192, 384] & [5, 5, 5] & 6 & [384, 192] & [1, 1] & 6\\
				\bottomrule	
			\end{tabular}
			\begin{tablenotes}
				\item[$\dag$]In addition to the 1024 input points, rendered images from different viewpoints of the point cloud are fed in.
			\end{tablenotes}	
		\end{threeparttable}
	}
	}	
\end{table*}

\subsection{Integrated Baselines}\label{sec3.4}
Our TPM is implemented based on existing point cloud self-supervised learning methods, including Point-MAE \cite{pang2022masked}, Point-M2AE \cite{zhang2022point}, Inter-MAE \cite{liu2023inter}, PointGPT-S \cite{chen2024pointgpt}, and PointGPT-B \cite{chen2024pointgpt}. For fair comparison, we do not modify any parameters of baseline methods except the number of masking tasks. The experimental settings involved are shown in Table \ref{tab1}.

The four baselines are introduced below, and more details can be obtained from the original articles.

\noindent\textbf{Point-MAE \cite{pang2022masked}.} A basic self-supervised mask learning scheme on point clouds that determines the theory and applicability of masks, patches, and autoencoder networks.

\noindent\textbf{Point-M2AE \cite{zhang2022point}.} A self-supervised approach that modifies the autoencoder into a pyramid architecture, progressively modeling spatial geometry to achieve hierarchical learning.

\noindent\textbf{Inter-MAE \cite{liu2023inter}.} Built on a point masking scheme, the image features after point cloud rendering are extracted to form an inter-modal comparison learning with the decoded features of the patched point patches.

\noindent\textbf{PointGPT-S \cite{chen2024pointgpt}.} Similar with Point-MAE, unordered point clouds are arranged into ordered sequences, and a dual masking strategy is used to predict point-wise patches.

\noindent\textbf{PointGPT-B \cite{chen2024pointgpt}.} Similar with PointGPT-S, except that 1) the pre-training dataset changes from ShapeNet \cite{chang2015shapenet} ($\sim$50k point clouds) to an unlabeled hybrid dataset (UHD), introducing 6 additional datasets \cite{song2015sun,mo2019partnet,uy2019revisiting,hackel2017semantic3d,wu20153d,armeni20163d} with a total of $\sim$300k point clouds and 2) the feature dimension double.

We summarize that these methods focus on data, modality, mask, and network. Thus, it is straightforward to show the strong applicability of our TPM. Indeed, the two mask rates we add are $m_1 = 0.5$ and $m_2 = 1-m_0$, where $m_1$ is to balance the potential confidence bias of the other two extreme masks and $m_2$ releases fine-grained completion signals.

\section{Experiments}
In this section, we first demonstrate the effectiveness of TPM in improving the baselines during pre-training. We then fine-tune and evaluate the SVM-guided pre-trained model by subjecting it to various downstream tasks. Finally, adequate ablation studies and analysis are performed to analyze the characteristics and principle of our TPM.

\subsection{Pre-training with TPM}\label{sec4.1}
\begin{table}[]
	\centering
	\caption{Linear SVM classification results (\%) on ModelNet40 \cite{wu20153d} and ScanObjectNN (PB\_T50\_RS) \cite{uy2019revisiting}. Different self-supervised learning methods are reported.}
	\label{tab2}
	\resizebox{0.9\linewidth}{!}{
		\begin{tabular}{lcc}
			\toprule
			Method &  ModelNet40 & PB\_T50\_RS \\
			\midrule
			3D-GAN \cite{wu2016learning} & 83.3 & - \\
			Latent-GAN \cite{achlioptas2018learning} & 85.7 & - \\
			SO-Net \cite{li2018so} & 87.3 & - \\
			FoldingNet \cite{yang2018foldingnet} & 88.4 & - \\
			VIP-GAN \cite{han2019view} & 90.2 & - \\
			\midrule
			DGCNN+Jagsaw3D \cite{sauder2019self} & 90.6 & 59.5 \\
			DGCNN+OcCo \cite{wang2021unsupervised} & 90.7 & 78.3 \\
			DGCNN+STRL \cite{qian2021spatiotemporal} & 90.9 & 77.9 \\
			DGCNN+CrossPoint \cite{afham2022crosspoint} & 91.2 & 81.7 \\
			DGCNN+CrossNet \cite{wu2023self} & 91.5 & 83.9 \\
			\midrule
			Point-BERT \cite{yu2022point} & 87.4 & - \\
			PM-MAE \cite{lin2024patchmixing} & 92.9 & - \\
			Point-MAE \cite{pang2022masked} & 90.8 & 77.1 \\
			\rowcolor{gray!15}Point-MAE + TPM & 91.2 (+0.4) & 78.2 (+1.1) \\
			Point-M2AE \cite{zhang2022point} & 92.9 & 83.1 \\
			\rowcolor{gray!15}Point-M2AE + TPM & \textbf{93.1} (+0.2) & \textbf{84.0} (+0.9)\\
			\bottomrule
	\end{tabular}}	
\end{table}

We pre-train Point-MAE and Point-M2AE with TPM on the ShapeNet \cite{chang2015shapenet}, which contains 57,448 object point clouds from 55 common categories. Additionally, the proposed method is compared with methods based on spatial reconstruction \cite{sauder2019self,wang2021unsupervised}, related data augmentation and transformation \cite{han2019view,qian2021spatiotemporal}, and contrastive learning \cite{afham2022crosspoint,wu2023self}.

\noindent\textbf{Linear SVM.} To evaluate the representational capabilities of the point cloud models generated during pre-training, we directly extract linear SVM features for the methods with our TPM on both synthetic ModelNet40 \cite{wu20153d} and real-world ScanObjectNN \cite{uy2019revisiting}. As shown in Table \ref{tab2}, for both classical PointMAE and PointGPT, TPM can fuel their discriminative capabilities, improving the accuracy by +0.4\%/+0.2\% and +1.1\%/+0.9\% on synthetic and real datasets, respectively. Experimental results show that TPM allows point clouds to discover more potential information during pre-training by modifying the masking task only.

\subsection{Fine-tuning with TPM}\label{sec4.2}
After pre-training, we discard all parameters in the $w_1$ and $w_2$ models as well as the decoder in $w_0$ and attach different network heads to the encoder in $w_0$. The new lightweight networks are fine-tuned to implement multiple downstream tasks at both the object level and scene level. 
\begin{table*}[htb]
	\centering
	\caption{Fine-tuned classification results (\%) on ScanObjectNN \cite{uy2019revisiting} and ModelNet40 \cite{wu20153d} datasets. Note that single-modal self-supervised methods only use point clouds as input, while cross-modal self-supervised methods introduce additional modal knowledge from pre-trained image models or generate additional modal knowledge. We do not make fair comparisons with them and only list them as references. Note that we evaluate three variants on the ScanObjectNN dataset and two types of point counts on the ModelNet40 dataset.}
	\label{tab3}
	\resizebox{0.9\linewidth}{!}{
		\begin{tabular}{lccccccc}
			\toprule 
			\multirow{2}{*}{Method} & \multirow{2}{*}{Reference} & \multirow{2}{*}{\#Parameters (M)} & \multicolumn{3}{c}{ScanObjectNN} & \multicolumn{2}{c}{ModelNet40} \\
			\cmidrule(lr){4-6} \cmidrule(lr){7-8}
			& & & OBJ\_BG & OBJ\_ONLY & PB\_T50\_RS & 1k & 8k \\
			\midrule
			\multicolumn{8}{c}{\textit{Supervised Learning Only}} \\
			\midrule
			PointNet \cite{qi2017pointnet} & CVPR 2017 & 3.5 & 73.3 & 79.2 & 68.0 & 89.2 & 90.8 \\
			PointCNN \cite{li2018pointcnn} & NeurIPS 2018 & 0.6 & 86.1 & 85.5 & 78.5 & 92.2 & - \\
			DGCNN \cite{wang2019dynamic} & TOG 2019 & 1.8 & 82.8 & 86.2 & 78.1 & 92.9 & - \\
			MVTN \cite{hamdi2021mvtn} & ICCV 2021 & 11.2 & 92.6 & 92.3 & 82.8 & 93.8 & - \\
			PointMLP \cite{ma2022rethinking} & ICLR 2022 & 12.6 & - & - & 85.4 & 94.5 & - \\
			PointNeXt \cite{qian2022pointnext} & NeurIPS 2022 & 1.4 & - & - & 87.7 & 94.0 & - \\
			\midrule
			\multicolumn{8}{c}{\textit{with  Single-Modal Self-Supervised Representation Learning}} \\
			\midrule
			Point-BERT \cite{yu2022point} & CVPR 2022 & 22.1 & 87.4 & 88.1 & 83.1 & 93.2 & 93.8 \\
			MaskPoint \cite{liu2022masked} & ECCV 2022 & 22.1 & 89.3 & 88.1 & 84.3 & 93.8 & - \\
			PM-MAE \cite{lin2024patchmixing} & TCSVT 2024 & 22.1 & 93.6 & 92.6 & 89.8 & 94.0 & - \\
			Point-MAE \cite{pang2022masked} & ECCV 2022 & 22.1 & 90.0 & 88.2 & 85.2 & 93.8 & 94.0 \\
			\rowcolor{gray!15}Point-MAE + TPM & - & 22.1 & 91.4 (+1.4) & 88.7 (+0.5) & 85.7 (+0.5) & 94.0 (+0.2) & 94.2 (+0.2) \\
			Inter-MAE \cite{liu2023inter} & TMM 2023 & 22.1 & 88.7 & 89.6 & 85.4 & 93.6 & 93.8 \\
			\rowcolor{gray!15}Inter-MAE + TPM & - & 22.1 & 91.0 (+2.3) & 90.4 (+0.8) & 85.6 (+0.2) & 94.0 (+0.4) & 94.0 (+0.2) \\
			Point-M2AE \cite{zhang2022point} & NeurIPS 2022 & 12.9 & 91.2 & 88.8 & 86.4 & 94.0 & - \\		
			\rowcolor{gray!15}Point-M2AE + TPM & - & 12.9 & 91.6 (+0.4) & 90.0 (+1.2) & 86.6 (+0.2) & 94.1 (+0.1) & - \\
			PointGPT-S \cite{chen2024pointgpt} & NeurIPS 2023 & 19.7 & 91.6 & 90.0 & 86.9 & 94.0 & 94.2 \\
			\rowcolor{gray!15}PointGPT-S + TPM & - & 19.7 & 91.8 (+0.2) & 89.8 (-0.2) & 86.8 (-0.1) & 93.8 (-0.2) & 94.1 (-0.1) \\
			PointGPT-B \cite{chen2024pointgpt} & NeurIPS 2023 & 82.6 & 95.8 & 95.2 & \textbf{91.9} & 94.4 & 94.6 \\
			\rowcolor{gray!15}PointGPT-B + TPM & - & 82.6 & \textbf{96.0} (+0.2) & \textbf{95.6} (+0.4) & 91.8 (-0.1) & \textbf{94.5} (+0.1) & \textbf{94.7} (+0.1) \\
			\midrule
			\multicolumn{8}{c}{\textit{with Cross-Modal Self-Supervised Representation Learning}}
			\\
			\midrule
			I2P-MAE \cite{zhang2023learning} & CVPR 2023 & 12.9 & 94.2 & 91.6 & 90.1 & 94.1 & - \\	
			TAP \cite{wang2023take} & ICCV 2023 & 12.6 & 90.4 & 89.5 & 85.7 & 94.0 & - \\
			ACT \cite{dong2022autoencoders} & ICLR 2023 & 22.1 & 93.9 & 91.9 & 88.2 & 93.7 & 94.0 \\		
			ReCon \cite{qi2023contrast} & ICML 2023 & 43.6 & 95.2 & 93.6 & 90.6 & 94.5 & 94.7 \\			 
			\bottomrule
		\end{tabular}
	}
\end{table*}
\begin{table}[]
	\centering
	\caption{Fine-tuned part segmentation results (\%) on ShapeNetPart \cite{yi2016scalable} dataset. The mean intersection over union (mIoU) of all classes (Cls.) and all instances (Ins.) is reported.}
	\label{tab4}
	\resizebox{0.8\linewidth}{!}{
		\begin{tabular}{lcc}
			\toprule 
			Method & Cls. mIoU & Ins. mIoU \\
			\midrule
			PointNet \cite{qi2017pointnet} & 80.4 & 83.7 \\
			PointCNN \cite{li2018pointcnn} & 84.6 & 86.1 \\
			DGCNN \cite{wang2019dynamic} & 82.3 & 85.2  \\
			PointMLP \cite{ma2022rethinking} & 84.6 & 86.1 \\
			\midrule
			Point-BERT \cite{yu2022point} & 84.1 & 85.6 \\
			MaskPoint \cite{liu2022masked}  & 84.4 & 86.0 \\
			PM-MAE \cite{lin2024patchmixing} & 84.3 & 85.9 \\
			Point-MAE \cite{pang2022masked} & 84.2 & 86.1 \\
			\rowcolor{gray!15}Point-MAE + TPM & 84.6 (+0.4) & 86.2 (+0.1) \\
			Inter-MAE \cite{liu2023inter} & 84.3 & 86.3 \\
			\rowcolor{gray!15}Inter-MAE + TPM & 84.6 (+0.3)  & 86.4 (+0.1) \\
			Point-M2AE \cite{zhang2022point} & 84.9 & 86.5 \\	
			\rowcolor{gray!15}Point-M2AE + TPM & 84.8 (-0.1) & 86.5 (+0.0) \\
			PointGPT-S \cite{chen2024pointgpt} & 84.1 & 86.2 \\
			\rowcolor{gray!15}PointGPT-S + TPM & 84.3 (+0.2) & 86.2 (+0.0) \\
			PointGPT-B \cite{chen2024pointgpt} & 84.5 & 86.5 \\
			\rowcolor{gray!15}PointGPT-B + TPM & \textbf{84.8} (+0.3) & \textbf{86.7} (+0.2) \\
			\midrule
			I2P-MAE \cite{zhang2023learning} & 85.2 & 86.8 \\	
			TAP \cite{wang2023take} & 85.2 & 86.9 \\
			ACT \cite{dong2022autoencoders} & 84.7 & 86.1 \\		
			ReCon \cite{qi2023contrast} & 84.8 & 86.4 \\			 
			\bottomrule
		\end{tabular}
	}		
\end{table}

\noindent\textbf{Object Classification.} We test the classification overall accuracy of the proposed method on both synthetic and real-world datasets. The selected pre-trained model is transferred to ScanObjectNN \cite{uy2019revisiting}, which contains about 15,000 objects (15 categories) extracted from real indoor scans, and ModelNet40 \cite{wu20153d}, which includes 12,311 clean 3D CAD objects (40 categories). For ScanObjectNN, we report three different experiments: OBJ-BG, OBJ-ONLY, and PB-T50-RS. For ModelNet40, in order to have a fair comparison, we use a standard voting strategy \cite{liu2019relation} for the tests, where the input point cloud contains only coordinate information.

The results in Table \ref{tab3} demonstrate that our TPM can bring an average +0.3\% improvement up to a maximum of 1.4\% in four baselines although it may change the original training and cause a few fluctuations. No additional parameter or component design is required to enable existing methods to achieve superior performance. Note that the improvement of TPM is more pronounced on ScanObjectNN than on ModelNet40. This phenomenon is in line with our expectation that the multi-mask task is designed to be useful for adapting to complex and comprehensive internal supervision, and is also directly reflected in complex objects.

\noindent\textbf{Part Segmentation.} We evaluate the impact of TPM for part segmentation on the ShapeNetPart \cite{yi2016scalable} dataset, which consists of 16,881 objects from 16 categories. For a fair comparison, we use the same segmentation head as in the baseline methods. Specifically, the input point cloud is sampled as 2048 points, and three hierarchical features at layers 4, 8 and 12 of the transformer blocks are extracted and concatenated. Subsequently, two features are obtained by average pooling, maximum pooling, concatenated and then up-sampling is executed to generate features for each point and MLP is applied for semantic prediction. The experimental results in Table \ref{tab4} demonstrate that our TPM provides significant positive enhancement for the part segmentation task that require fine-grained representations. 

\noindent\textbf{Few-shot Learning.} To demonstrate the generalizability of TPM on few-shot learning, we conduct experiments on the repartitioned ModelNet40 dataset. Following previous work \cite{pang2022masked,zhang2022point,chen2024pointgpt}, there are four different setups using the $w$-way, $s$-shot paradigm. Specifically, $w$ denotes the number of randomly selected classes and $s$ denotes the number of sampled objects per selected class. The results are shown in Table \ref{tab5}, where our TPM shows that it exhibits incremental effects in all tests. Especially for PointGPT-B, due to the pre-training process with massive data and multiple masks, the few-shot learning in the fine-tuning process basically predicts various objects and creates a state-of-the-art performance close to 100\%. This demonstrates the ability of TPM to power existing methods to acquire generalized knowledge even under the constraints of low data.
\begin{table*}[]
	\centering
	\caption{Fine-tuned few-shot classification results (\%) on ModelNet40 \cite{wu20153d} dataset. Ten independent experiments are performed, and the mean accuracy ($\uparrow$) and standard deviation ($\downarrow$) are reported.}
	\label{tab5}
	\resizebox{0.7\linewidth}{!}{	
		\begin{tabular}{lcccc}
			\toprule
			Method & 5-way 10-shot & 5-way 20-shot & 10-way 10-shot & 10-way 20-shot \\
			\midrule 
			DGCNN \cite{wang2021unsupervised}& 91.8$\pm$3.7 & 93.4$\pm$3.2 & 86.3$\pm$6.2 & 90.9$\pm$5.1 \\ 
			OcCo \cite{wang2021unsupervised}& 91.9$\pm$3.3 & 93.9$\pm$3.1 & 86.4$\pm$5.4 & 91.3$\pm$4.6 \\
			CrossPoint \cite{afham2022crosspoint} & {92.5$\pm$3.0} & {94.9$\pm$2.1} & {83.6$\pm$5.3} & {87.9$\pm$4.2} \\
			\midrule 
			Point-BERT \cite{yu2022point} & 94.6$\pm$3.1 & 96.3$\pm$2.7 & 91.0$\pm$5.4 & 92.7$\pm$5.1 \\
			MaskPoint \cite{liu2022masked} & 95.0$\pm$3.7 & 97.2$\pm$1.7 & 91.4$\pm$4.0 & 93.4$\pm$3.5 \\
			PM-MAE \cite{lin2024patchmixing} & 96.7$\pm$2.7 & 97.6$\pm$1.6 & 92.6$\pm$4.6 & 95.3$\pm$3.5 \\
			Point-MAE \cite{pang2022masked} & 96.3$\pm$2.5 & 97.8$\pm$1.8& 92.6$\pm$4.1& 95.0$\pm$3.0\\
			\rowcolor{gray!15}Point-MAE + TPM & 96.6$\pm$2.5 & 97.4$\pm$2.1 & \textbf{93.7$\pm$4.1} & 95.2$\pm$3.2 \\
			Inter-MAE \cite{liu2023inter} & 95.3$\pm$2.1 & 97.7$\pm$1.4& 91.2$\pm$3.7 & 94.0$\pm$3.8 \\
			\rowcolor{gray!15}Inter-MAE + TPM & 97.0$\pm$1.9 & 97.6$\pm$1.6& 93.0$\pm$4.6 & 95.1$\pm$2.8 \\
			Point-M2AE \cite{zhang2022point} & 96.8$\pm$1.8 & 98.3$\pm$1.4 & 92.3$\pm$4.5 & 95.0$\pm$3.0 \\
			\rowcolor{gray!15}Point-M2AE + TPM & 96.5$\pm$1.9 & 97.9$\pm$1.8 & 92.4$\pm$4.4 & 95.4$\pm$3.1 \\
			PointGPT-S \cite{chen2024pointgpt} & 96.8$\pm$2.0 & 98.6$\pm$1.1 & 92.6$\pm$4.6 & 95.2$\pm$3.4\\
			\rowcolor{gray!15}PointGPT-S + TPM & 97.0$\pm$2.1 & 98.6$\pm$1.3 & 92.7$\pm$4.8 & 95.3$\pm$3.7 \\
			PointGPT-B \cite{chen2024pointgpt} & 97.5$\pm$2.0 & 98.8$\pm$1.0 & 93.5$\pm$4.0 & \textbf{95.8$\pm$3.0} \\
			\rowcolor{gray!15}PointGPT-B + TPM & \textbf{97.7$\pm$1.6} & \textbf{98.8$\pm$0.8} & 93.3$\pm$4.1 & 95.8$\pm$3.2 \\
			\midrule
			I2P-MAE \cite{zhang2023learning} & 97.0$\pm$1.8 & 98.3$\pm$1.3 & 92.6$\pm$5.0 & 95.5$\pm$3.0 \\	
			TAP \cite{wang2023take} & 97.3$\pm$1.8  & 97.8$\pm$1.7 & 93.1$\pm$2.6 & 95.8$\pm$1.0 \\	
			ACT \cite{dong2022autoencoders} &96.8$\pm$2.3  & 98.0$\pm$1.4 & 93.3$\pm$4.0 & 95.6$\pm$2.8 \\		
			ReCon \cite{qi2023contrast} & 97.3$\pm$1.9 & 98.9$\pm$1.2 & 93.3$\pm$3.9 & 95.8$\pm$3.0 \\	
			\bottomrule
		\end{tabular}
	}		
\end{table*}
\begin{table}[]
	\centering
	\caption{Ablation study: mask construction for pre-training.}
	\label{tab6}
	\begin{tabular}{lcc}
		\toprule
		Mask construction & ModelNet40 & PB\_T50\_RS \\
		\midrule
		0.6 $\rightarrow$[0.6, 0.4] & 93.8 & 85.6 \\
		0.6 $\rightarrow$[0.7, 0.3] & 93.6 & 85.4 \\
		0.6 $\rightarrow$[0.8, 0.2] & 93.2 & 85.1 \\
		0.6 $\rightarrow$[0.9, 0.1] & 92.8 & 84.6 \\
		\midrule
		\rowcolor{gray!15}0.6 $\rightarrow$[0.6, 0.5, 0.4] & \textbf{94.0} & \textbf{85.7} \\
		0.6 $\rightarrow$[0.7, 0.5, 0.3] & 93.5 & 84.8 \\
		0.6 $\rightarrow$[0.8, 0.5, 0.2] & 93.6 & 85.2 \\
		\midrule
		0.6 $\rightarrow$[0.7, 0.6, 0.5, 0.4] & 93.4 & 85.2 \\
		0.6 $\rightarrow$[0.6, 0.5, 0.4, 0.3] & 93.2 & 84.7 \\
		\bottomrule
	\end{tabular}
\end{table}
\begin{table}[t]
	\centering
	\caption{Ablation study: loss construction for pre-training.}
	\label{tab7}
	\begin{tabular}{lcc}
		\toprule
		Loss construction & ModelNet40 & PB\_T50\_RS \\
		\midrule
		$\mathcal{\lambda}_{m_i}=1$ & 93.8 & 85.4 \\
		\rowcolor{gray!15}$\mathcal{\lambda}_{m_i}=m_i/{sum(\{m_i\})}$ & \textbf{94.0} & \textbf{85.7} \\
		\bottomrule
	\end{tabular}
\end{table}
\begin{figure*}[]
	\centering
	\includegraphics[width=0.9\linewidth]{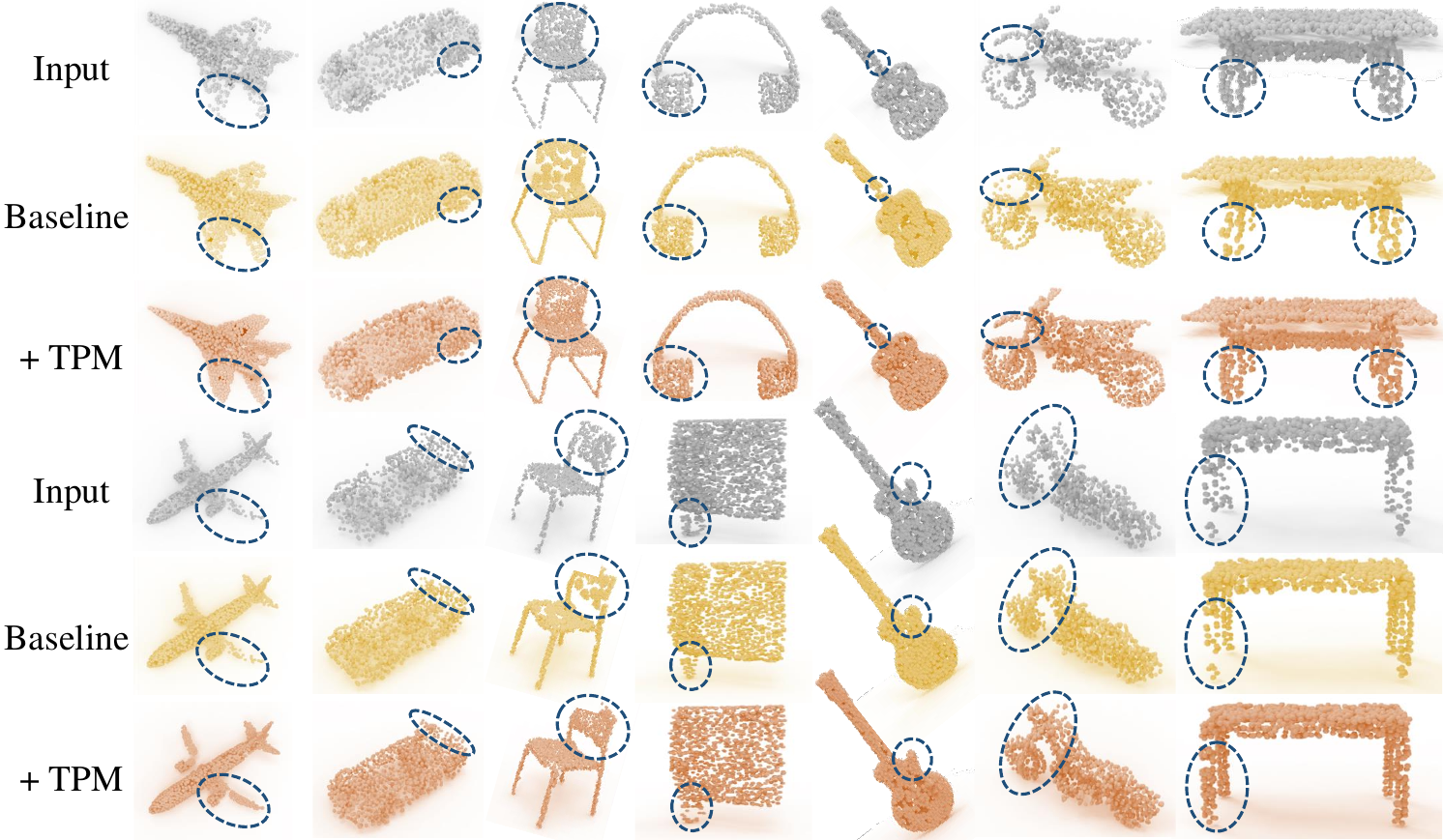}
	\caption{Completion visualization of the baseline with/without TPM on the ShapeNet \cite{chang2015shapenet} dataset. Our TPM focuses more on detail areas.}
	\label{fig4}
\end{figure*}
\begin{table}[t]
	\centering
	\caption{Ablation study: weight selection for fine-tuning.}
	\label{tab8}
	\begin{tabular}{lcc}
		\toprule
		Weight selection &  ModelNet40 & PB\_T50\_RS \\
		\midrule
		\rowcolor{gray!15}$w_0 \mapsto m_0$ & \textbf{94.0} & \textbf{85.7} \\
		$w_1 \mapsto m_1$ & 93.6 & 85.4 \\
		$w_2 \mapsto m_2$ & 93.0 & 85.1 \\
		$w_{0,1,2} \mapsto m_0|m_1|m_2$ & 93.0 & 85.1 \\
		\bottomrule
	\end{tabular}
\end{table}
\subsection{Ablation Study for TPM}\label{sec4.3}	
Since our core contribution is the introduction of TPM, there is no need to consider the strengths or weaknesses present in existing methods. As a result, we conduct ablation studies on pre-trained mask and loss construction and fine-tuned weight selection with “Point-MAE + TPM”. We assess the impact of these designs by reporting the object classification accuracy achieved by the fine-tuned model on the ModelNet40 (1k) and ScanObjectNN (PB\_T50\_RS).

\noindent\textbf{Mask construction.} We notice that when only a small mask is added, a certain spatial awareness is also produced in the pre-training process, and it is even possible to surpass triple point masking with carefully designed binary point mask. However, we still reveal that the triple point masking plays a stable completion role, and $m_0,m_1,m_2=[0.6, 0.5, 0.4]$ derived from $m_0=0.6$ is the most suitable configuration, see Table \ref{tab6}. If the number of masks increases again, the completion task in pre-training is overloaded, making it difficult to parse the completion of different masks.

\noindent\textbf{Loss construction.} Due to the triple masks provided, there are triple completions during pre-training. In order to balance the different levels of completion, we set the loss weights $\mathcal{\lambda}_{m_i}=m_i/{sum(\{m_i\})}$ that are proportional to the mask values for the point patches to be completed. Table \ref{tab7} shows that the mask-based loss weights can enhance TPM to generate discriminative attention. Although the effect is only a small improvement over setting the same weights for different masks, due to the fact that the addition of triple masks already contributes considerably to the effectiveness of self-supervised learning, this provides more reliable principles for potentially more masks and more masking approaches.

\noindent\textbf{Weight selection.} Since we adopt the triple point maskinging strategy, and the linear SVM needs to evaluate the different performances of the same model facing different situations. That is, $w_{0,1,2}$ acts on each mask from each epoch in the pre-training process. Through theoretical analysis and experimental results, our choice is $w_0 \mapsto m_0$ due to taking into account two factors: 1) pre-training sets a more difficult pretext task to better serve downstream tasks, and 2) SVM's guidance is to measure the distinguishability of the completed point cloud. Therefore, we observe in Table \ref{tab8} that the fourth weight selection ($w_{0,1,2} \mapsto m_0|m_1|m_2$) is often the same as the easiest task ($w_2 \mapsto m_2$) and can easily achieve high linear classification effects. In contrast, this selection cannot be adapted to downstream tasks.

\begin{figure}[]
	\centering
	\includegraphics[width=\linewidth]{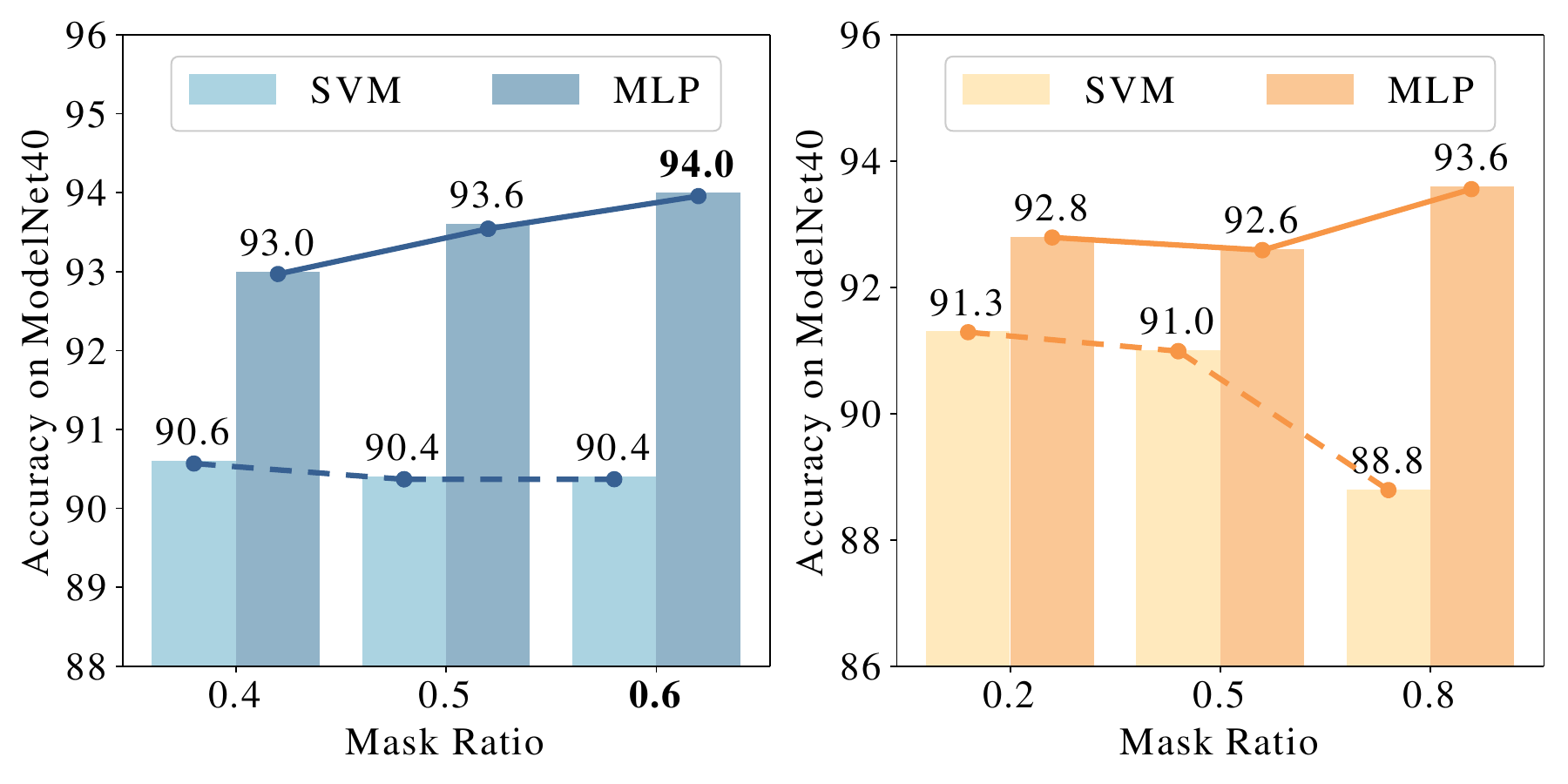}
	\caption{Comparison of SVM classification during pre-training and MLP prediction during fine-tuning under different masks.}
	\label{fig5}
\end{figure}
\subsection{More Analysis for TPM}\label{sec4.4}
We first illustrate the facilitation of TPM during pre-training by comparing the completed examples with and without it, as shown in Figure \ref{fig4}. It can be found that the baselines with TPM have better robustness and realism for completed parts. Then the mechanism and specific performance of masks and weights are further analyzed in depth.

\noindent\textbf{Mask analysis.} We show the results under [0.6, 0.5, 0.4] and [0.8, 0.5, 0.2] mask constructions in Figure \ref{fig5}, including SVM classification during pre-training and MLP prediction during fine-tuning. We argue that in the PointMAE setting, the simple pretext task with a low mask (\ie, 0.2) does not result in an effective gain for the downstream task. This suggests that TPMs constructed based on baseline masks are suitable for the existing baseline and that weight selection needs to consider both SVM and task difficulty.

\begin{figure}[]
	\centering
	\includegraphics[width=\linewidth]{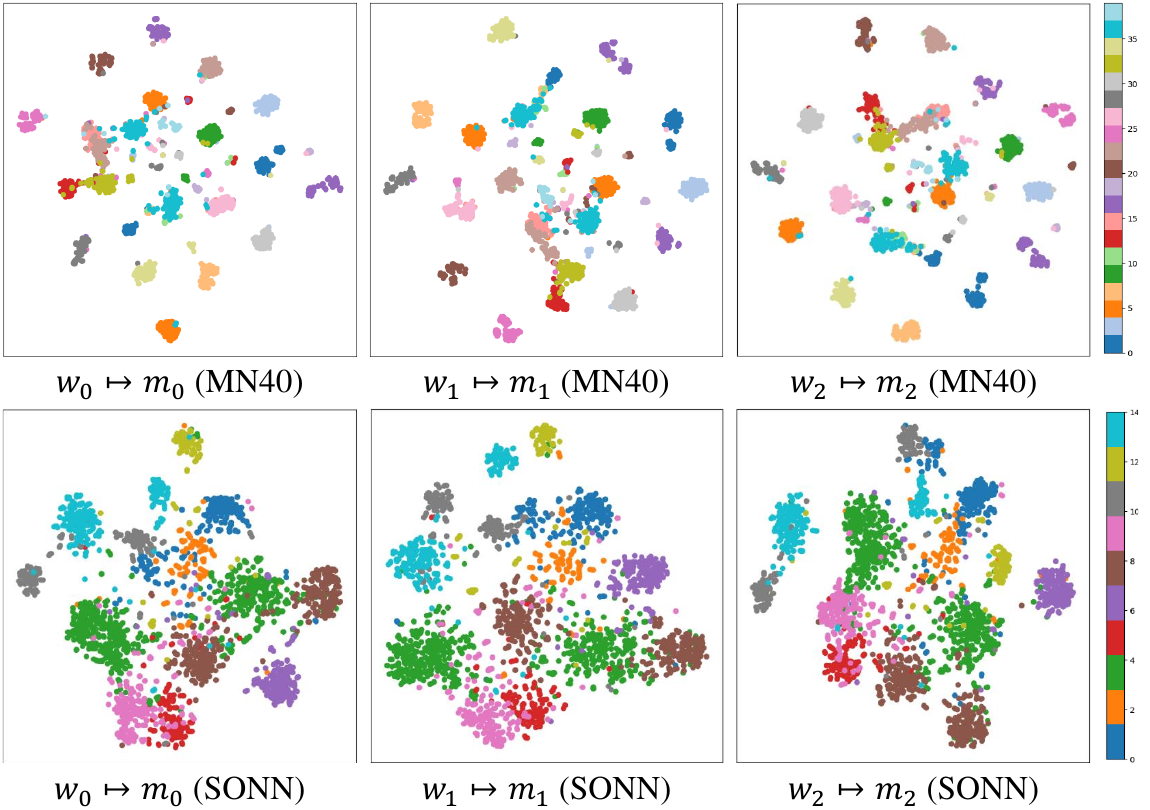}
	\caption{Visualization of feature discrimination with optimal model weights under different masks during the fine-tuning stage, implemented by t-SNE \cite{van2008visualizing} on the MN40 and SONN datasets.}
	\label{fig6}
\end{figure}

\noindent\textbf{Weight analysis.} To further illustrate the effective role of triple masks, we explain this phenomenon by analyzing the optimal weights they produce. Specifically, we maintain the optimal weights of triple masks by TPM and perform fine-tuning experiments at ModelNet40 (MN40) and ScanObjectNN (SONN). As reflected in Figure \ref{fig6}, even though a particular model of TPM has difficulty in distinguishing features with similar semantic labels under the mask at that time, this distinction may be “meetable” in models under other masks. Thus, this fine-grained semantic discrimination enhances the learning capability of triple masks.

\noindent\textbf{Limitation.} TPM is undoubtedly a straightforward and effective technique for self-supervised learning on point clouds. Nevertheless, it is difficult to find a universal mask construction due to the different masking and completion ways from baselines. In theory, there are variable combinations of masks and networks. Moreover, we particularly show that abundant data is benificial to promote self-supervised learning \cite{chen2024pointgpt}, and our TPM can amplify this advantage.

\section{Conclusion}
In this paper, we propose TPM, an effective and scalable multi-masking scheme that addresses the domain gap between generative and downstream tasks for 3D self-supervised learning. Diverging from conventional 3D mask modeling methods, TPM systematically enriches the shape perception of 3D objects through well-designed triple point masking. The use of SVM-guided weight selection strategy in the pre-trained models augments its discriminative reliability on new tasks. Results suggest that our TPM yields noteworthy improvements over unimodal self-supervised methods without the need for cross-modal information.

\bibliographystyle{IEEEtran}
\bibliography{IEEEabrv,reference}

\vfill

\end{document}